\documentclass{article} % For LaTeX2e
\usepackage{iclr2021_conference,times}

\usepackage{amsmath,amsfonts,bm}

\def\eqref#1{equation~\ref{#1}}
\def\1{\bm{1}}

\def\vc{{\bm{c}}}

\def\vx{{\bm{x}}}

\def\vz{{\bm{z}}}

\DeclareMathAlphabet{\mathsfit}{\encodingdefault}{\sfdefault}{m}{sl}
\SetMathAlphabet{\mathsfit}{bold}{\encodingdefault}{\sfdefault}{bx}{n}

\def\sS{{\mathbb{S}}}

\newcommand{\E}{\mathbb{E}}

\usepackage[utf8]{inputenc} % allow utf-8 input
\usepackage[T1]{fontenc}    % use 8-bit T1 fonts
\usepackage{hyperref}       % hyperlinks
\usepackage{url}            % simple URL typesetting
\usepackage{booktabs}       % professional-quality tables
\usepackage{amsfonts}       % blackboard math symbols
\usepackage{nicefrac}       % compact symbols for 1/2, etc.
\usepackage{microtype}      % microtypography

\usepackage{epigraph}
\usepackage{enumitem}
\usepackage{wrapfig}
\usepackage{graphicx}
\usepackage{subfigure}
\usepackage{amsmath}
\usepackage{multirow}
\usepackage{tabularx}
\usepackage{wrapfig}
\usepackage{lipsum}
\usepackage{array}
\usepackage{mathtools}
\usepackage{icomma}
\usepackage{siunitx}
\usepackage{csquotes}
\usepackage{xspace}
\usepackage[subtle,tracking=normal]{savetrees}

\usepackage{subfigure}
\usepackage{amsmath,amsthm,amssymb}
\newcommand\cut[1]{}

\newcolumntype{C}[1]{>{\centering\arraybackslash}m{#1}}
\newcolumntype{R}[1]{>{\raggedleft\arraybackslash}m{#1}}

\newcommand{\be}{\begin{equation}}
\newcommand{\ee}{\end{equation}}
\newcommand{\bea}{\begin{eqnarray}}
\newcommand{\eea}{\end{eqnarray}}
\newcommand{\beaa}{\begin{eqnarray*}}
\newcommand{\eeaa}{\end{eqnarray*}}

\DeclareMathAlphabet{\mathpzc}{OT1}{pzc}{m}{n}

\newcommand{\betavae}{\ensuremath{\beta}-VAE\xspace}

\usepackage{color}
\newboolean{draftmode}
\setboolean{draftmode}{false}

\newcommand{\mycomment}[3]{{\textcolor{#3}{[#1 #2]}}}
\newcommand{\ihmarker}{{\textcolor{blue}{\ensuremath{^{\textsc{I}}_{\textsc{H}}}}}}
\newcommand{\almarker}{{\textcolor{red}{\ensuremath{^{\textsc{A}}_{\textsc{L}}}}}}
\newcommand{\lmmarker}{{\textcolor{purple}{\ensuremath{^{\textsc{L}}_{\textsc{M}}}}}}
\newcommand{\nwmarker}{{\textcolor{olive}{\ensuremath{^{\textsc{N}}_{\textsc{W}}}}}}
\newcommand{\cbmarker}{{\textcolor{teal}{\ensuremath{^{\textsc{C}}_{\textsc{B}}}}}}
\newcommand{\sdmarker}{{\textcolor{magenta}{\ensuremath{^{\textsc{S}}_{\textsc{D}}}}}}

\ifthenelse{\boolean{draftmode}}{
 \newcommand{\ih}[1]{\mycomment{\ihmarker}{#1}{blue}}
 \newcommand{\al}[1]{\mycomment{\almarker}{#1}{red}}
 \newcommand{\lm}[1]{\mycomment{\lmmarker}{#1}{purple}} 
 \newcommand{\nw}[1]{\mycomment{\nwmarker}{#1}{olive}} 
 \newcommand{\cb}[1]{\mycomment{\cbmarker}{#1}{teal}} 
 \newcommand{\sd}[1]{\mycomment{\sdmarker}{#1}{magenta}}
}{
 \newcommand{\ih}[1]{}
 \newcommand{\al}[1]{}
 \newcommand{\lm}[1]{}
 \newcommand{\nw}[1]{}
 \newcommand{\cb}[1]{}
 \newcommand{\sd}[1]{}
}

\title{Disentangling Action Sequences: Finding Correlated images}

\author{Jiantao Wu,  Lin Wang\textsuperscript{*} \\
Shandong Provincial Key Laboratory of Network Based Intelligent Computing\\
University of Jinan\\
Jinan, 250022, China \\
\texttt{\{clouderow,wangplanet\}@gmail.com} 
}

\iclrfinalcopy % Uncomment for camera-ready version, but NOT for submission.
\begin{document}

\maketitle

\begin{abstract}
  Disentanglement is a highly desirable property of representation due to its similarity with human’s understanding and reasoning. This improves interpretability, enables the performance of down-stream tasks, and enables controllable generative models. However, this domain is challenged by the abstract notion and incomplete theories to support unsupervised disentanglement learning. We demonstrate the data itself, such as the orientation of images, plays a crucial role in disentanglement and instead of the factors, and the disentangled representations align the latent variables with the action sequences. We further introduce the concept of disentangling action sequences which facilitates the description of the behaviours of the existing disentangling approaches. An analogy for this process is to discover the commonality between the things and categorizing them. 

  Furthermore, we analyze the inductive biases on the data and find that the latent information thresholds are correlated with the significance of the actions. For the supervised and unsupervised settings, we respectively introduce two methods to measure the thresholds. We further propose a novel framework, fractional variational autoencoder (FVAE), to disentangle the action sequences with different significance step-by-step. Experimental results on dSprites and 3D Chairs show that FVAE improves the stability of disentanglement.
\end{abstract}

\section{Introduction}
The basis of artificial intelligence is to understand and reason about the world based on a limited set of observations.
Unsupervised disentanglement learning is highly desirable due to its similarity with the way we as human thinking.
For instance, we can infer the movement of a running ball based on a single glance  this is because the human brain is capable of disentangling the positions from a set of images.
It has been suggested that a disentangled representation is helpful for a large variety of downstream tasks \citep{Schlkopf2012OnCA,Peters2017ElementsOC}.
According to \cite{Kim.2018},  a disentangled representation promotes interpretable semantic information.
That brings substantial advancement, including but not limited to reducing the performance gap between humans and AI approaches \citep{Lake.2017,Higgins2018SCANLH}.
Other instances of disentangled representation include semantic image understanding and generation \citep{Lample.2017, Zhu.2018GANs, Elgammal.2017}, zero-shot learning \citep{Zhu.2019}, and reinforcement learning \citep{HigginsPRMBPBBL17}.
Despite the advantageous of the disentangling representation approaches, there are still two issues to be addressed including the abstract notion and the weak explanations.

\paragraph{Notion} the conception of disentangling factors of variation is first proposed in 2013.
It is claimed in \cite{Bengio.2013} that for observations the considered factors should be explanatory and independent of each other.
The explanatory factors are however hard to formalize and measure.
An alternative way is to disentangle the ground-truth factors \citep{Ridgeway2016ASO,Do.2020}.
However, if we consider the uniqueness of the ground-truth factors, a question arise here is how to discover it from multiple equivalent representations? 
As a proverb ``one cannot make bricks without straw'', \cite{Locatello.2019} prove the impossibility of disentangling factors without the help of inductive biases in the unsupervised setting.

\paragraph{Explanation} There are mainly two types of explanations for unsupervised disentanglement: information bottleneck, and independence assumption.
The ground-truth factors affect the data independently, therefore, the disentangled representations must follow the same structure.
The approaches, holding the independence assumption, encourage independence between the latent variables  \citep{Schmidhuber.1992,Chen.2018,Kim.2018,Kumar2018DIPVAE,Lopez2018InformationCO}.
However, the real-world problems have no strict constraint on the independence assumption, and the factors may be correlative.
The other explanation incorporates information theory into disentanglement.
\citeauthor{Burgess.2018,Higgins.2017,InsuJeon.2019,Saxe.2018} suggest that a limit on the capacity of the latent information channel promotes disentanglement by enforcing the model to acquire the most significant latent representation.
They further hypothesize that the information bottleneck enforces the model to find the significant improvement.
% However, they didn't explore the reasons causing different contributions to the reconstruction loss. related work里介绍

% disentangle action sequence 
In this paper, we first demonstrate that instead of the ground-truth factors the disentangling approaches learn actions of translating based on the orientation of the images.
We then propose the concept of disentangling actions which discover the commonalities between the images and categorizes them into sequences.
We treat disentangling action sequences as a necessary step toward disentangling factors, which can capture the internal relationships between the data, and make it possible to analyze the inductive biases from the data perspective.
Furthermore, the results on a toy example show that the significance of actions is positively correlated with the threshold of latent information.
Then, we promote that conclusion to complex problems.
Our contributions are summarized in the following:
\begin{itemize}
    \item We show that the significance of action is related to the capacity of learned latent information, resulting in the different thresholds of factors.
    \item We propose a novel framework, fractional variational autoencoder (FVAE) to extracts explanatory action sequences step-by-step, and at each step, it learns specific actions by blocking others’ information.
\end{itemize}

We organize the rest of this paper as follows.
Sec.2 describes the development of unsupervised disentanglement learning and the proposed methods based on VAEs.
In Sec.3, through an example, we show that the disentangled representations are relative to the data itself and further introduce a novel concept, i.e., disentangling action sequences.
Then, we investigate the inductive biases on the data and find that the significant action has a high threshold of latent information.
In Sec.4, we propose a step-by-step disentangling framework, namely fractional VAE (FVAE), to disentangle action sequences.
For the labelled and unlabelled tasks, we respectively introduce two methods to measure their thresholds.
We then evaluate FVAE on a labelled dataset (dSprites, \cite{dsprites17}) and an unlabelled dataset (3D Chairs, \cite{Aubry2014Chairs}).
Finally, we conclude the paper and discuss the future work in Sec.5
\section{Unsupervised disentanglement learning}
We first introduce the abstract concepts and the basic definitions, followed by the explanations based on information theory and other related works. 
This article focuses on the explanation of information theory and the proposed models based on VAEs.  
\subsection{The concept}
Disentanglement learning is fascinating and challenging because of its intrinsic similarity to human intelligence.
As depicted in the seminal paper by \citeauthor{Bengio.2013}, humans can understand and reason from a complex observation to the explanatory factors.
A common modelling assumption of disentanglement learning is that the observed data is generated by a set of ground-truth factors.
Usually, the data has a high number of dimensions, hence it is hard to understand, whereas the factors have a low number of dimensions thus simpler and easier to be understood.
The task of disentanglement learning is to uncover the ground-truth factors, such factors are invisible to the training process in an unsupervised setting.
The invisibility of factors makes it hard to define and measure disentanglement \citep{Do.2020}.

Furthermore, it is shown in \cite{Locatello.2019} that it is impossible to unsupervisedly disentangle the underlying factors for the arbitrary generative models without inductive biases.
In particular, they suggest that the inductive biases on the models and the data should be exploited.
However, they do not provide a formal definition of the inductive bias and such a definition is still unavailable.
% TODO 目前的方法多从因子的角度。因子之间是独立的，latent structure is same as factor's. 通过隐变量学习因子的特性实现解耦。
% 研究数据本身对解耦的影响。如图像的形状，朝向。发现样本之间的关联就是解耦。

\subsection{Information bottleneck}
Most of the dominant disentangling approaches are the variants of variational autoencoder (VAE).
The variational autoencoder (VAE) is a popular generative model, assuming that the latent variables obey a specific prior (normal distribution in practice).
The key idea of VAE is maximizing the likelihood objective by the following approximation:

\begin{equation}\label{eq:vae}
    \mathcal{L}(\theta, \phi; x,z) = \E_{q_\phi(\mathbf{z}|\mathbf{x})}[\log{p_\theta (x|z)}] - 
   D_{\mathrm{KL}}(q_\phi(z|x) || p(z)),
\end{equation}

which is know as the evidence lower bound (ELBO); 
where the conditional probability $P(x|z),Q(z|x)$ are parameterized with deep neural networks.

\citeauthor{Higgins.2017} find that the KL term of VAEs encourages disentanglement and introduce a hyperparameter $\beta$ in front of the KL term.
They propose the \betavae maximizing the following expression:
\begin{equation}\label{eq:beta-vae}
\mathcal{L}(\theta, \phi; x,z) = \E_{q_\phi(\mathbf{z}|\mathbf{x})}[\log{p_\theta (x|z)}] - 
 \beta D_{\mathrm{KL}}(q_\phi(z|x) || p(z)).
\end{equation}
$\beta$ controls the pressure for the posterior $Q_{\phi}(z|x)$ to match the factorized unit Gaussian prior $p(z)$.
Higher values of $\beta$ lead to lower implicit capacity of the latent information and ambiguous reconstructions.
\citeauthor{Burgess.2018} propose the Annealed-VAE that progressively increases the information capacity of the latent code while training:
\begin{equation}\label{eq:annealed-vae}
\mathcal{L}(\theta, \phi; x,z,C) = \E_{q_\phi(\mathbf{z}|\mathbf{x})}[\log{p_\theta (x|z)}] - 
 \gamma \left| D_{\mathrm{KL}}(q_\phi(z|x) || p(z)) - C \right|
\end{equation}
where $\gamma$ is a large enough constant to constrain the latent information, $C$ is a value gradually increased from zero to a large number to produce high reconstruction quality.
As the total information bottleneck gradually increasing, they hypothesize that the model will allocate the capacity of the most improvement of the reconstruction log-likelihood to the encoding axes of corresponding factor.
% TODO
% xxx prove this hypothesis by PCA theory.
However, they did not exploit why each factor makes the different contributions to the reconstruction log-likelihood.

\subsection{Other related work}
The other dominant direction  initiates from the prior of factors.
They assume that the ground-truth factors are independent of each other, and a series of methods enforce the latent variables have the same structure as the factors.
FactorVAE \citep{Kim.2018} applies a discriminator to approximately calculate the total correlation (TC, \cite{Watanabe.1960});
$\beta$-TCVAE \citep{Chen.2018} promotes the TC penelty by decomposing the KL term;
DIP-VAE \citep{Kumar2018DIPVAE} identifies the covariance matrix of $q(z)$.
% \citep{Rolinek2019VariationalAP} explains the process of disentangling as a PCA-like behavior. 
However, the real-world problems have no strict constrain in the prior of factors.
For instance, the length of hair and the gender are two independent factors to describe a person, but the real situation is that women are more likely to have long hair.
% \cite{Higgins.2017} promotes the variational posterior $q(z|x)$ to be similar to a factorial $p(z)$.
% \cite{Kim.2018,Chen.2018} improve total correlation to be similar to the prior $p(z)$.
% Other technics like the covariance matrix of q(z) (Kumar et al., 2017), a kernel-based measure of independence (Lopez et al., 2018), also benefits the independence assumption.

\begin{figure}[tbp]
\centering
\includegraphics[width=0.8\linewidth]{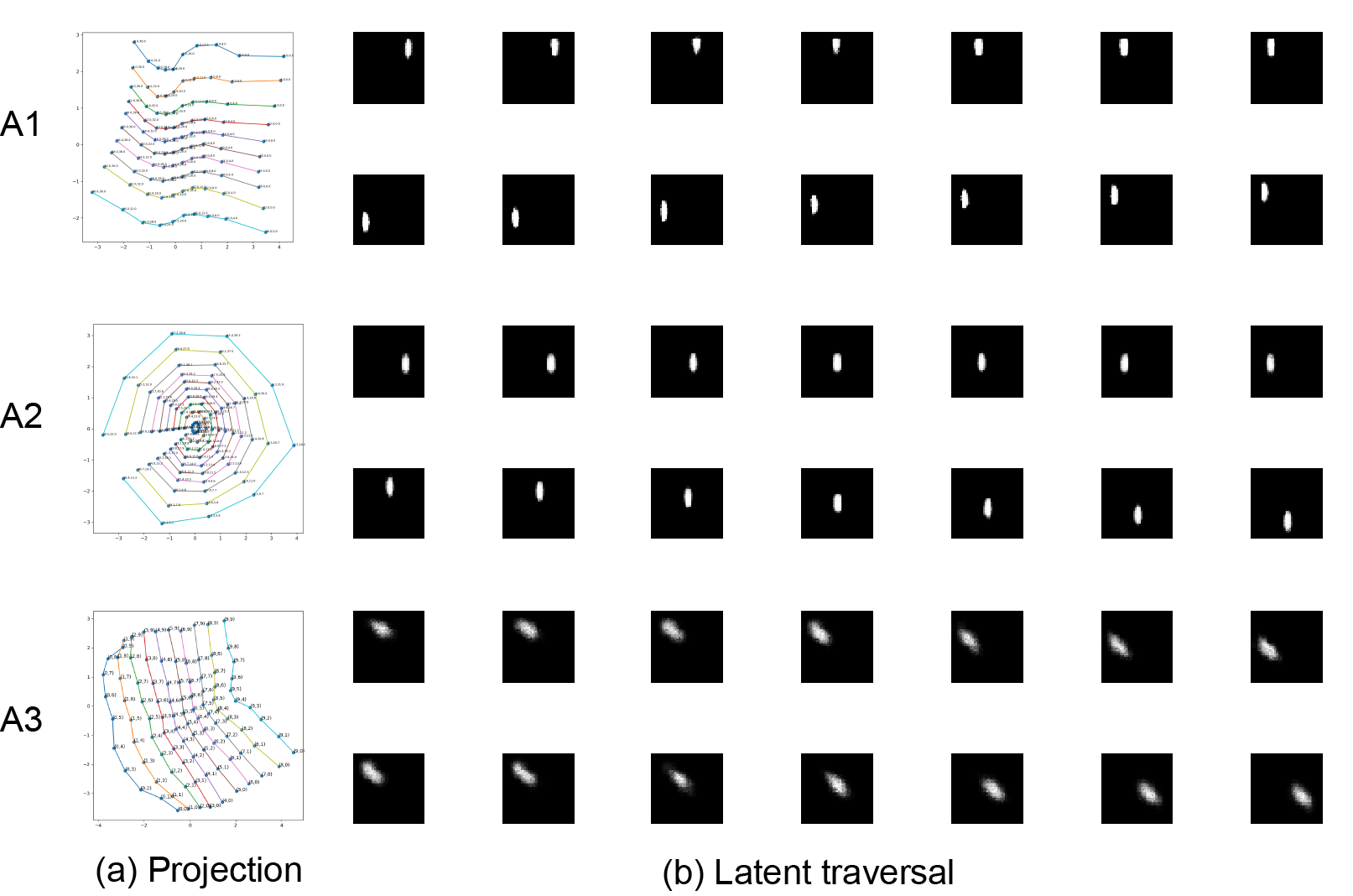}
\caption{Leaned presentations on A1, A2, and A3. (a) The factors are projected into the latent space. (b) Each row denotes the latent traversal on the specific dataset.}
\label{fig:leaned_presentation}
\end{figure}

\section{Disentangling action sequences}

For machines and humans, disentangling the underlying factors is a challenging task.
For instance, there are more than 1,400 breeds of dogs in the world, and it seems impossible for an ordinary person to distinguish all of them just by looking at their pictures.
The challenge in disentangling the underlying factors is mainly due to the complexity of establishing relationships without supervision, where the corresponding models should contain some level of prior knowledge or inductive biases.
However, it is possible to determine the differences without having extensive knowledge.
For example, one may mistakenly identify the breed of a dog by looking at a picture but is almost impossible to misrecognize a dog as a cat.
Therefore, in practice, discovering the differences or similarities is often a much easier task than that of uncovering the underlying factors, and it does not also need much prior knowledge.
One may conclude that discovering the commonalities between the things is an important step toward disentanglement.

\subsection{Disentangled representations}
Although the disentangling approaches can learn informative, independent, and meaningful representations on dSprites \citep{dsprites17}, these models disentangle the ground-truth factors by accident.
In other words, there is no reason for the latent variables to have any structure since there always exists a decoder that can map the latent variable into the desired output.
To determine the representations that the current approaches learn, here we design a dataset family to reveal the common aspects of these representations.

We create a toy dataset family--- each dataset contains 40x40 images which are generated from an original image of a 11x5 rectangle, by translating on a 64x64 canvas.
Each image on the dataset has a unique label (position of X, position of Y) to describe the ground-truth factors.
In this dataset family, there are two variables: the orientation of the rectangle and the way to determine the two factors.
There are infinite solutions to determine these two factors; the polar coordinate system and the Cartesian coordinate system are the most common solution.
We then create a baseline dataset, A1, with a horizontal rectangle in the Cartesian coordinate system and obtain its variants.
A2 differs in the positions determined by the polar coordinate system, and A3 differs in the orientation (45 degrees) of the images.
For the experiment settings, we choose the well-examined baseline model, \betavae ($\beta=50$), and the backbone network follows the settings in \cite{Locatello.2019}.

As it is shown in Fig. 1(a), we visualize the learned representation on the latent space.
In the left column, each point represents an image with the annotation of the corresponding factors.
We also link the points with the same controlling factor with coloured lines to show the ground-truth transformation.
We argue that the factors are not the key to disentanglement since the learned representations are changed while the factors are unchanged (A1, A3), and the learned representations do not change while the factors are changed (A1, A2).
From Fig. \ref{fig:leaned_presentation}, one can see that the invariant in these three cases is that they learn a representation moving along the direction of the long side of the rectangle and the orthogonality direction.
The disentangled representations consider the rectangle as the reference system.

\subsection{Action sequences }
We assume that images of the observed data consist of a set of actions such as rotation, translation, and scaling.
Given a subset of images from original data set, we define the \textbf{meaningful action} as a sequence, i.e. an ordered permutation of elements from the subset.
To obtain such sequences, one can get an action by traversing the ground-truth factors:
\begin{equation}
    \sS=\{\vx_i\}=T(\{\vc_i\}),
\end{equation}
where $T$ is a transformation, $\vc_i$ is the set of factors, $\vx_i$ is an image from the dataset.
Therefore, the disentanglement learning can be interpreted as a process of learning meaningful action sequences.
Furthermore, these actions have associated parameters to illustrate the changing processes in detail, i.e., a parameterized action sequence represents an action.
We define an action sequence as a subset of images from the dataset, and it reveals the relationships amongst the images.
The formal definition of action sequences can be described as the latent traversal in \cite{Higgins.2017}.

\begin{figure}
    \centering
    \includegraphics[width=0.9\linewidth]{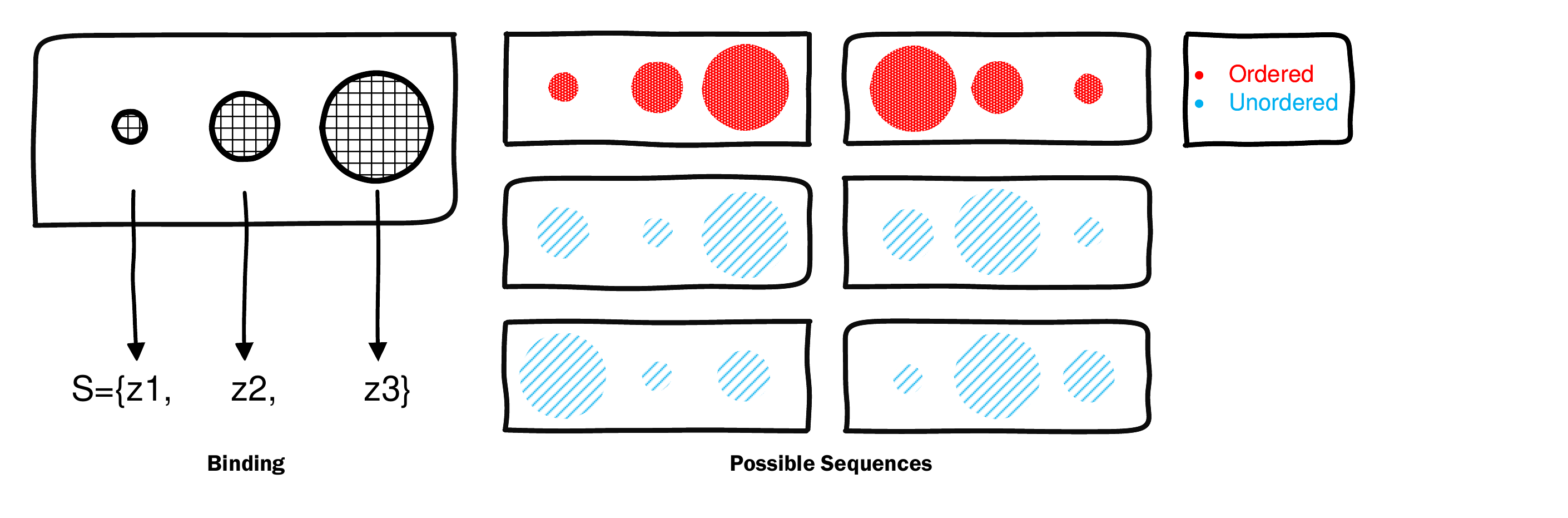}
    \caption{An example of action sequences. There are three images with differing sizes on the dataset (left box), and the AE learns a representation with one dimension of latent space (the double-arrow lines). So the total number of possible sequences is six. Two of them are meaningful, and the others are somewhat random.}
    \label{fig:seqs}
\end{figure}

For an autoencoder (AE) architecture, the neural networks are flexible enough to approximate any actions.
For clarity, we define an action as a sequence of images controlled by the ground-truth factors, and an action sequence is the approximation of this action generated by a neural network.
% The neural network plays a critical role in organizing images from the dataset into packages of sequences.
Due to the strong expression of neural networks, the sequences may be organized in any given order.
As shown in \ref{fig:seqs}, the neural network can form any possible sequences, and among them, the ordered sequences follow human’s intuition---the circle becomes larger or smaller.
Using VAEs, minimizing the objective increase the overlap between the posterior distributions across the dataset \cite{Burgess.2018}, and it then leads to disentanglement and emerges human intuition.

\begin{figure}[tbp]
\centering
    \subfigure[Entropy]{
    \includegraphics[width=0.45\textwidth]{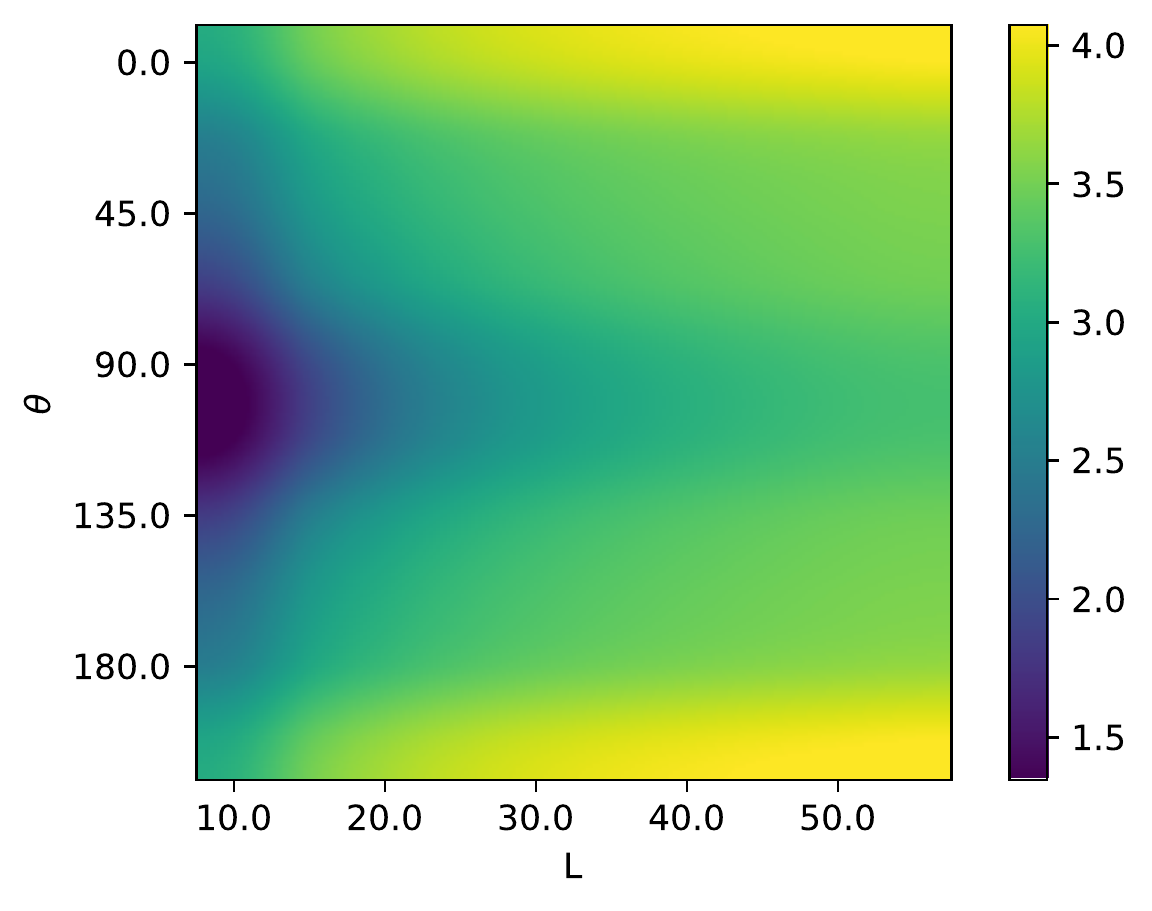}
    }
    \subfigure[KL divergence]{
    \includegraphics[width=0.45\textwidth]{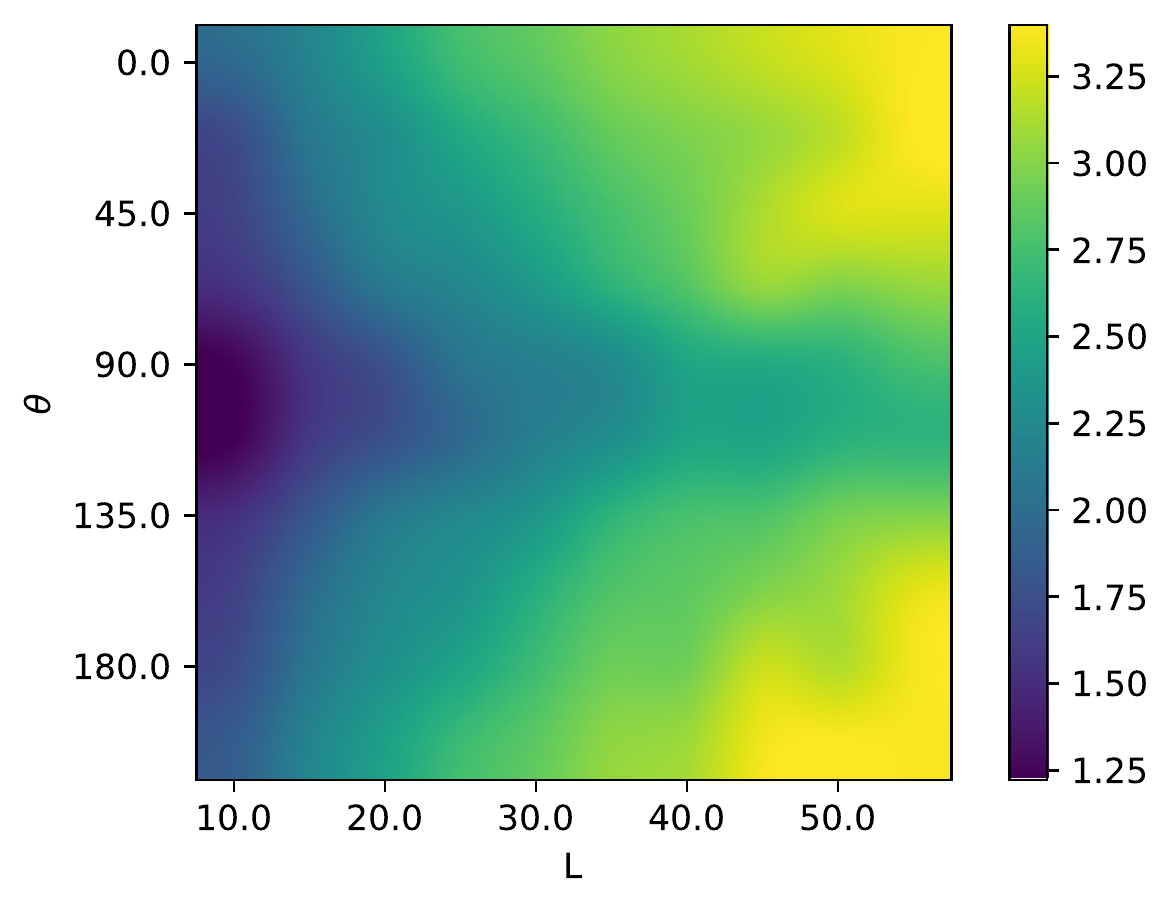}
    }
\centering
\caption{Comparison of entropy and KL divergence on A4.}
\label{fig:H_vs_KL}
\end{figure}

\subsection{Significance of sequences}
Although \cite{Locatello.2019} addresses the importance of inductive biases, few studies investigate the biases for disentanglement on the data.
As we have shown in Sec. 3.1, the orientation of the rectangle can affect the direction of disentangled representation.
We assume that there are much more inductive biases on the data which have not been yet exploited.
It is suggested in \cite{Burgess.2018} that the latent components have different contributions to the objective function.
Here, we investigate the reasons for different contributions and show that the capacity of latent information is correlated to the significance of the sequences.
To measure the significance of actions, we use the information among this action:
\begin{equation}
    \displaystyle H(\sS) = -\int_\sS p(\vx) \mathrm{log} (p(\vx))d\vx,
\end{equation}
where $\sS$ is the action controlled by the factor, .
% TODO 

We design a translation family (A4) with two controllable parameters, $\theta, L$ to determine the significance of the sequence; $\theta$ is the orientation of the rectangle; $L$ is the maximal distance of translating.
All datasets have only one factor which is the position of $X$-axis, and also contain one sequence.
If the value of $L$ is small enough, the images in the sequence are almost the same, this sequence is referred to as insignificant.
Both $\theta, L$ have different effects on the significance of the sequence.
As shown in Fig. \ref{fig:H_vs_KL}, the trend of KL divergence is consistent with that of entropy. Please note that the maximum for both are reached when $\theta=90$ and $L$ is at its maximum. These imply that the significance is closely related to information among the sequence.

% TODO 不够详细
Therefore, a higher significance of the sequence results in more information on the latent variables.
In contrast, by the gradual increase of the pressure on the KL term, the latent information decreases until it reaches zero.
One can infer that there exists a threshold of latent information.
Experimental results show a positive correlation between the significance of data and the threshold of latent information.

\section{Plain implementation according to thresholds}
\begin{figure}[ht!]
    \subfigure[FVAE]{\includegraphics[width=0.7\linewidth]{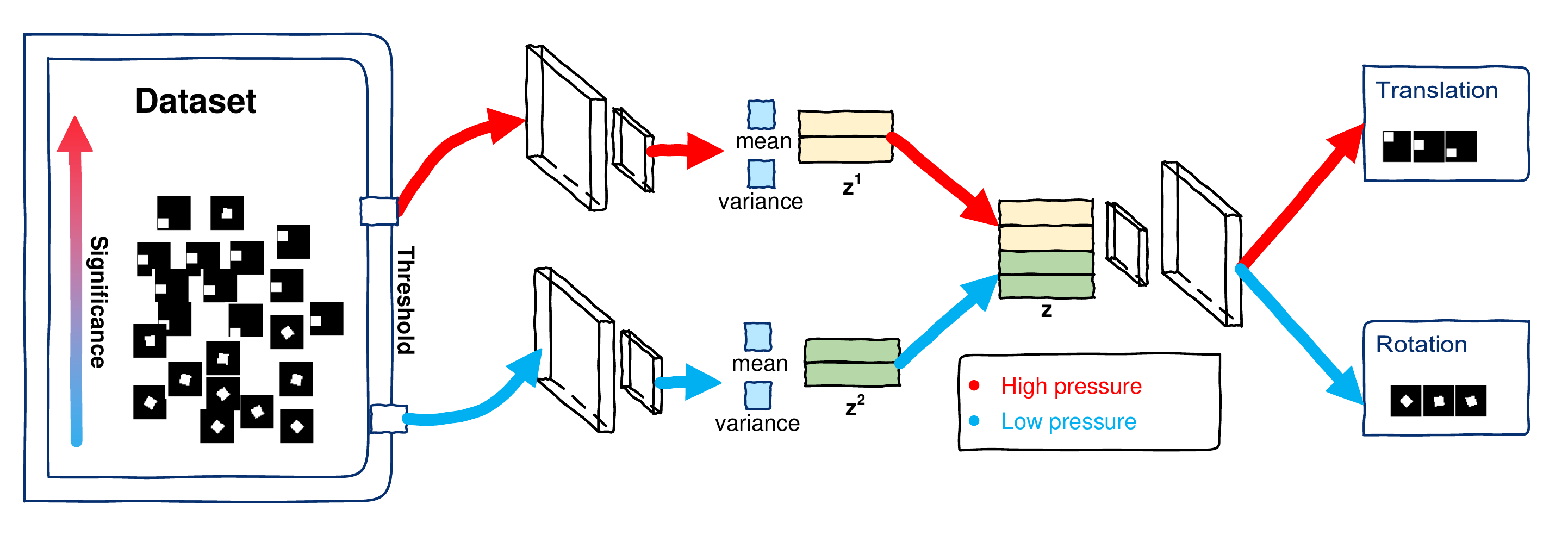}
    }%
    \subfigure[Mixed label]{\includegraphics[width=.28\linewidth]{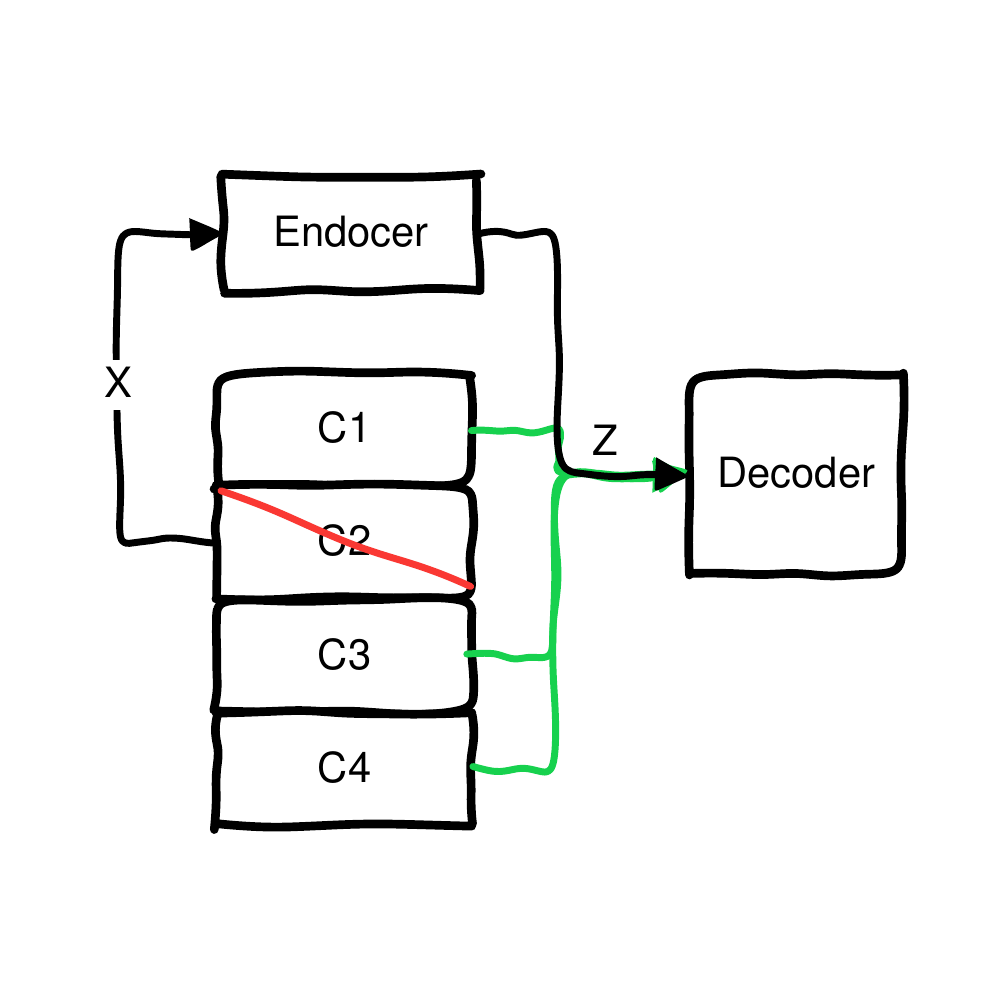}
    }
    \caption{(a) The architecture of FVAE. Though the samples distribute in the dataset randomly, they have intrinsic significance. Under a high pressure ($\beta$), the significant actions can pass information alone the red line to itself along, and the information of insignificant actions is blocked. (b) The decoder receive the label information except for the target action.}
    \label{fig:architecture}
\end{figure}

We have discussed the situation with only one action and there are various thresholds for different significant actions.
Particularly, if $\beta$ is large enough, information of the insignificant actions will be blocked, leading to the disentangling process decays to a single factor discovery problem.
% TODO
From the modelling perspective, the learning process is similar to a single action learning problem.
However, the difficulty of disentanglement is that different kinds of ground-truth actions are mixed, and a single and a fixed parameter $\beta$ is unable to separate them.
Therefore, a plain idea is to set different thresholds on the learning phases, and then in each phase, we enforce the model to learn specific actions by blocking information of the secondarily significant actions.
We propose a fractional variational autoencoder (FVAE) which disentangles the action sequences step-by-step.
The architecture of FVAE is shown in Fig. 4(a).
The encoder consists of several groups of sub-encoders, and the inputs of the decoder are the concatenated codes of all sub-encoders.
Besides, to prevent re-entangling the learned actions, we set different learning rates for the sub-encoders, which reduces the learning rate for the reset of N-1 groups and prevent the model from allocating the targeted action into the leaned codes.
The training process of FVAE is similar to the common operation in chemistry for separating mixtures---distillation.
To separate a mixture of liquid, we repeat the process of heating the liquid with different boiling points, and in each step, the heating temperature is different to ensure that only one component is being collected.

\textbf{Discussion}
Although AnnealedVAE follows the same principles as the FVAE, it differs in the interpretation of the effects of beta, and it does not explicitly prevent mixing the factors.
Moreover, the performance of AnnealedVAE depends on the choice of hyperparameter in practice \cite{Locatello.2019}.
``A large enough value'' is hard to determine, and the disentangled representation is re-entangled for an extremely large C.
To address this issue, here we introduce two methods to determine the thresholds in each phase for the labelled and unlabelled tasks.

\begin{figure}
    \centering
    \subfigure[]{
        \includegraphics[width=.45\linewidth]{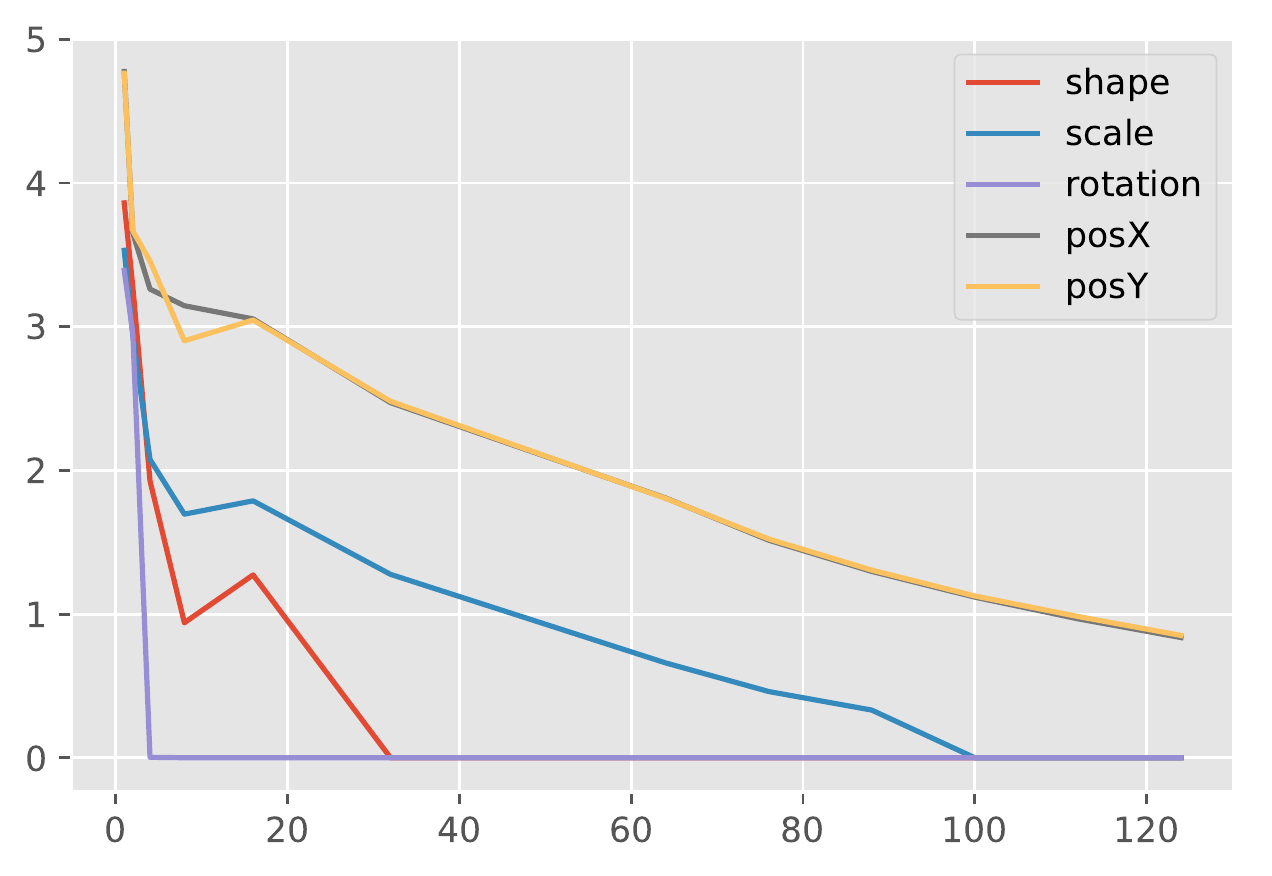}
        \label{fig:threshold_dsprites}
    }
    \subfigure[]{
        \includegraphics[width=.45\linewidth]{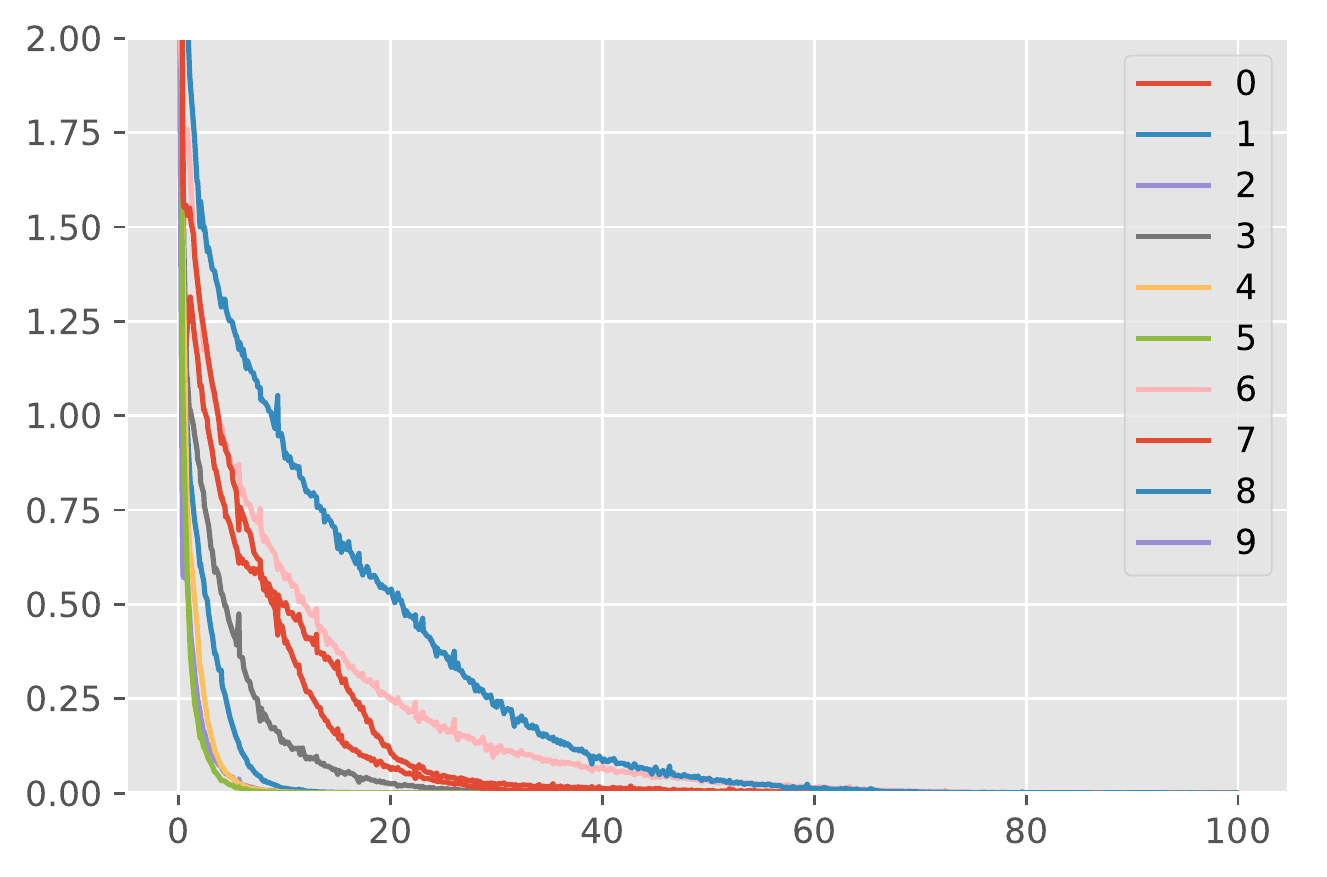}
        \label{fig:threshold_chairs}
    }
    
    \caption{$\beta$ vs KL divergence on dSprites (left) and 3D Chairs (right). }
    \label{fig:threshold}
\end{figure}

\begin{figure}
    \centering
    \subfigure[\betavae]{
        \includegraphics[width=.23\linewidth]{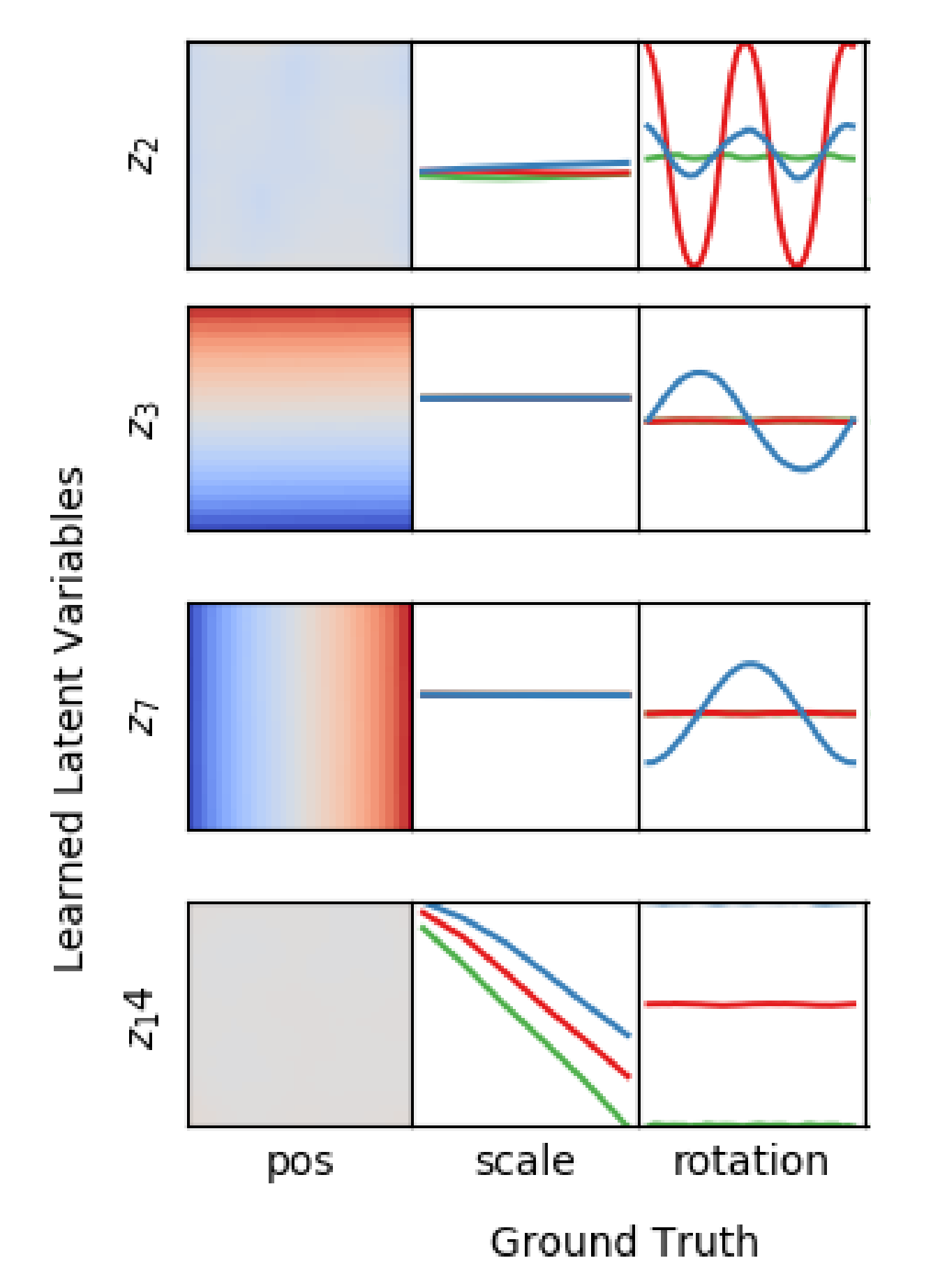}
    }%
    \subfigure[FVAE]{
        \includegraphics[width=.23\linewidth]{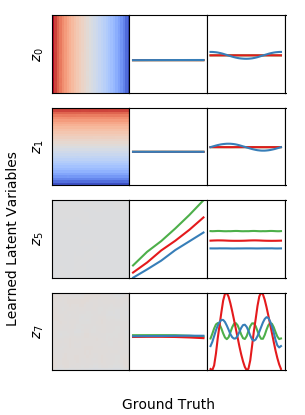}
    }%
    \subfigure[Disentanglement]{
        \includegraphics[width=.48\linewidth]{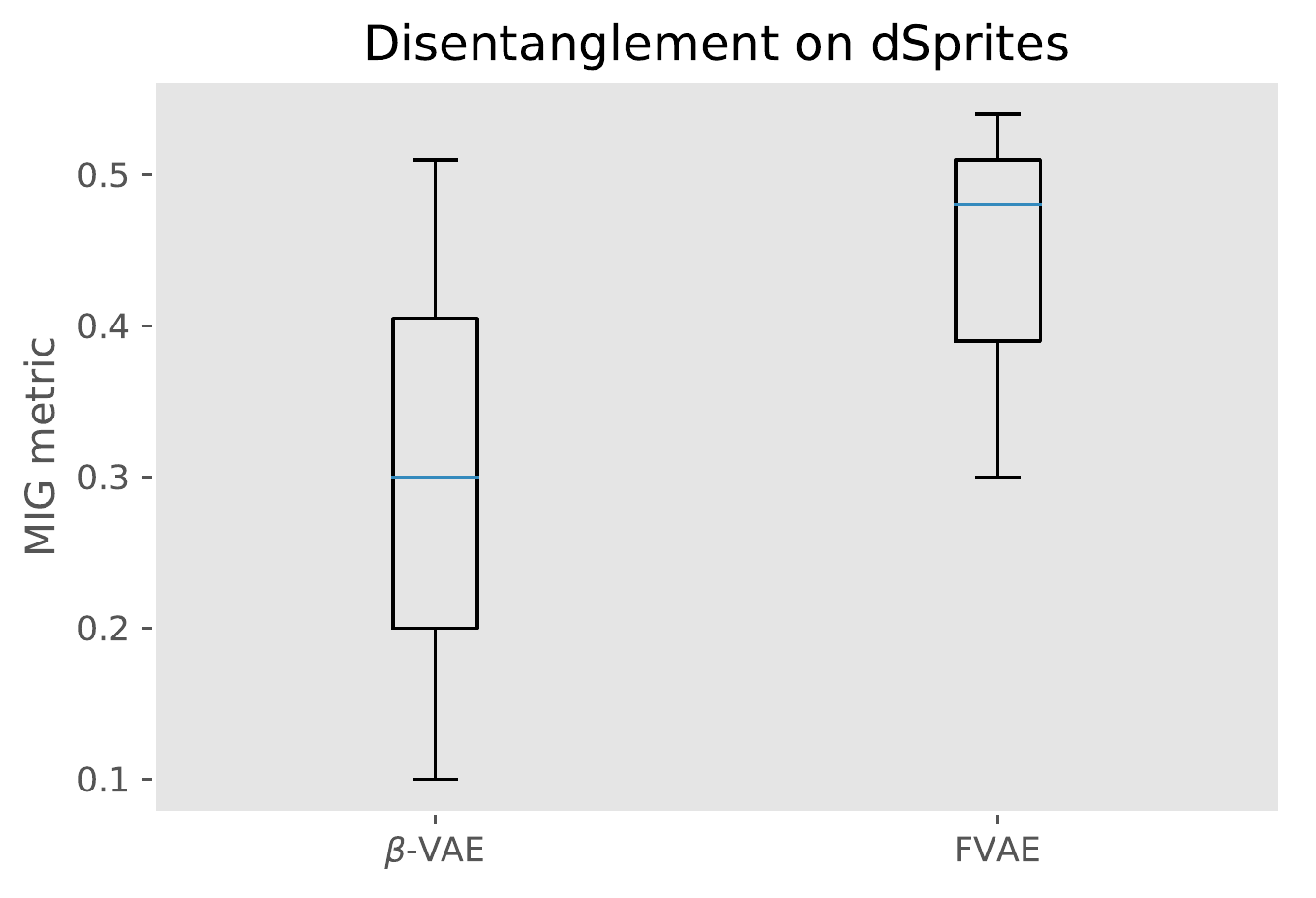}
    }
    \caption{Comparison between \betavae and FVAE. (a) and (b) show the relationship between $\vz$ and factors on dSprites (only active units are shown). (c) The disentanglement scores of \betavae and FVAE (MIG, \cite{Chen.2018}). 
}
    \label{fig:comparison}
\end{figure}

\subsection{Labelled task}
% TODO rewrite
%得先阐述一下问题。我们已经在简单问题（单个action）上证明了the significance of actions is positive correlated to the threshold.是对于复杂问题：包含各类actions还没有验证。所以使用了人工数据集dsprites，来验证以上结论的推广能力.最直接的办法是对某类action遍历所有可能的sequence。比如对于平移X来说，共有32x40x6x3种不同的action。这种方式的计算成本太大。所以我们提出一种混合标签信息的训练方式。

For the labelled setting, we focus on one type of actions and clip the rest of them at first. 
However, the samples of one action are usually insufficient.
For example, there are only three types of shapes on dSprites.
Besides, the label information may be corrupted and only some parts of the dataset are labelled.

To address these issues, we introduce the architecture shown in Fig. 4(b), in which the label information excepted for the target actions are directly provided to the decoder.
We evaluate FVAE on dSprites dataset (involving five actions: translating, rotating, scaling, and shaping).
We first measure the threshold of each action, and the result is shown in Fig. \ref{fig:threshold_dsprites}.
One can see that the thresholds of translating and scaling are higher than the others.
This suggests that these actions are significant and easy to be disentangled.
This is in line with the results in  \cite{Burgess.2018,Higgins.2017}.

According to these thresholds, we then arrange three stages for dSprites.
At each stage, we set up a bigger number of $\beta$ than the threshold of the secondary action.
The pressure on the KL term also prevents the insignificant actions from being disentangled and ensures that the model only learns from the information of the target action.
The training of each stage can be found in the Appendix.
As shown in Fig. 7(a), the translation factor is disentangled at first easily, while it is hard to distinguish the shape, orientation, and scale. 
Gradually, scaling and orientation also emerge in order. 
Nevertheless, it should be noted that the shaping is still hard to be separated. 
This could be attributed to the lack of commonalities between these three shapes on dSprites and motion compensation for a smooth transition. 
% TODO 运动补偿,https://zh.wikipedia.org/wiki/%E8%BF%90%E5%8A%A8%E8%A1%A5%E5%81%BF
In other words, in terms of shape, the lack of intermediate states between different shapes is an inevitable hurdle for its disentanglement.
Fig. \ref{fig:comparison} shows more substantial difference between the \betavae and the FVAE. 
\betavae has an unstable performance compared to FVAE, and position information entangles with orientation on some dimension.

\subsection{Unlabelled task}
For the unlabelled setting, we introduce the annealing test to detect the potential components.
In the beginning, a very large value for $\beta$ is set to ensure that no action is learned.
Then, we gradually decrease $\beta$ to disentangle the significant actions.
There exists a critical point in which the latent information starts increasing and that point approximates the threshold of the corresponding action.

3D Chairs is an unlabelled dataset containing 1394 3D models from the Internet.
Fig. \ref{fig:threshold_chairs} shows the result of the annealing test on 3D Chairs.
One can recognize three points where the latent information suddenly increases: 60, 20, 4.
Therefore, we arrange a three-stage training process for 3D Chairs (more details in the Appendix).
As shown in Fig. 7(b), one can see the change of azimuth in the first stage.
In the second stage, one can see the change of size and in the third stage, one can see the change of the leg style, backrest, and material used in the making of the chair.
\begin{figure*}
    \centering
    \subfigure[dSprites]{
    \begin{minipage}[]{0.42\linewidth}
        \includegraphics[width=\linewidth]{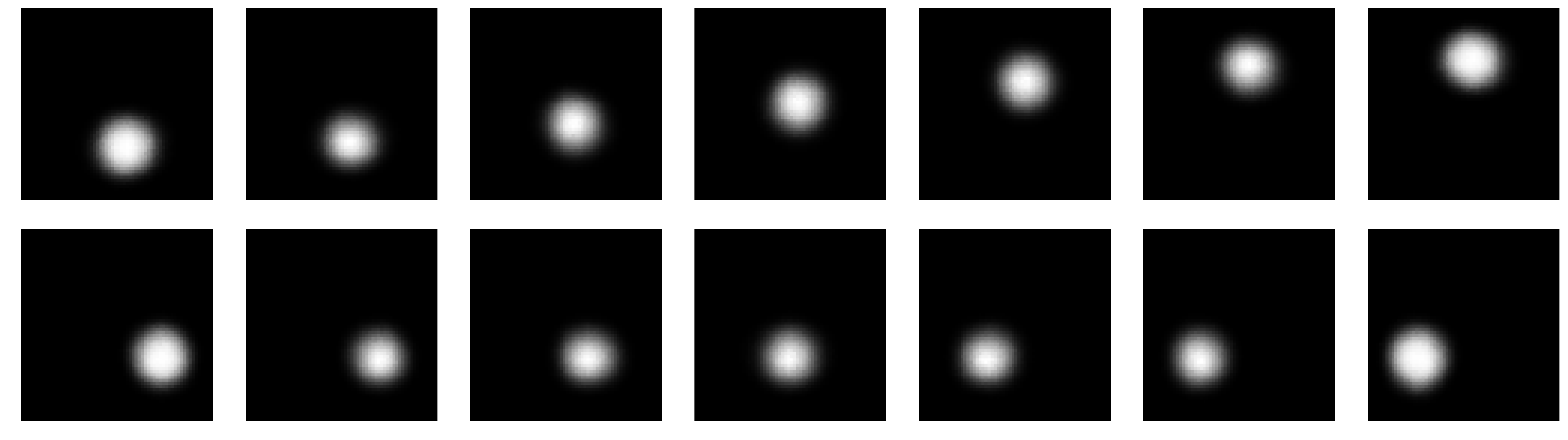}\\ 
        \centering\small{Stage (1)}\\
        \includegraphics[width=\linewidth]{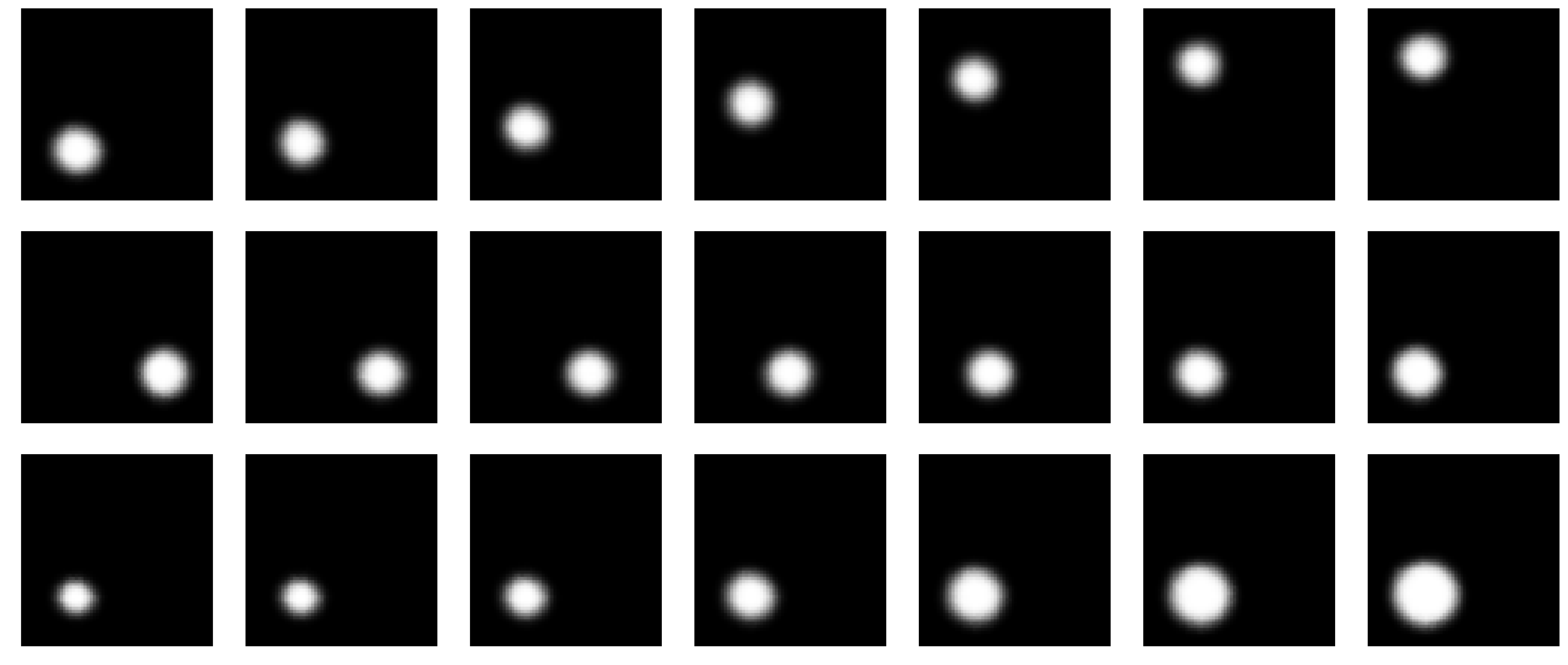}\\
        \centering\small{Stage (2)}\\
        \includegraphics[width=\linewidth]{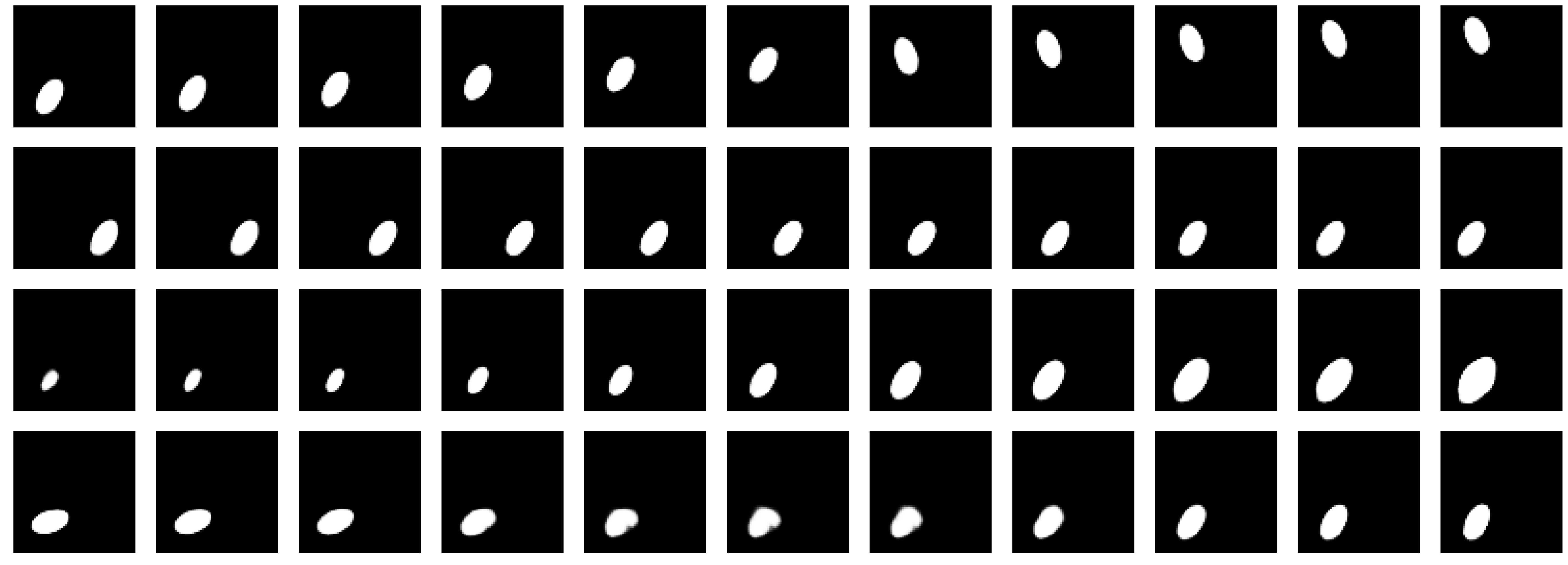}
        \centering\small{Stage (3)}\\
    \end{minipage}%
    }%
    \hspace{0.05\linewidth}
    \subfigure[3D Chairs]{
    \begin{minipage}[]{0.5\linewidth}
        
        \includegraphics[width=\linewidth]{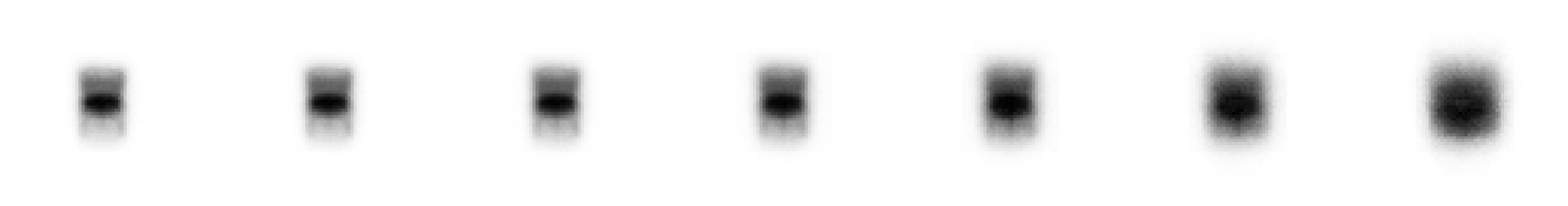}\\
        \centering\small{Stage (1)}\\
        \includegraphics[width=\linewidth]{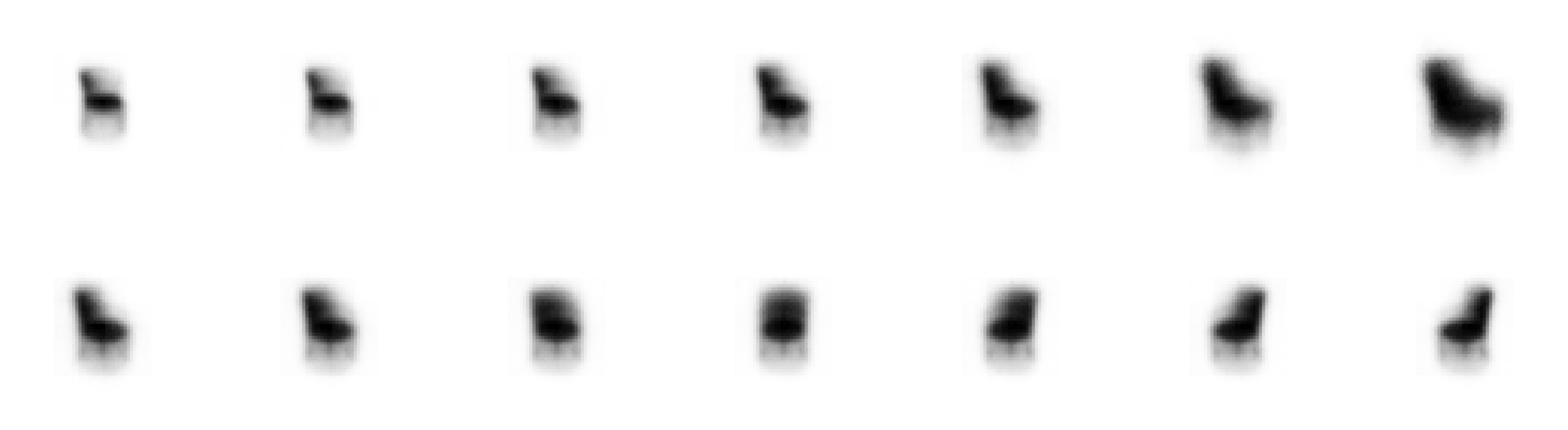}\\
        \centering\small{Stage (2)}\\
        \includegraphics[width=\linewidth]{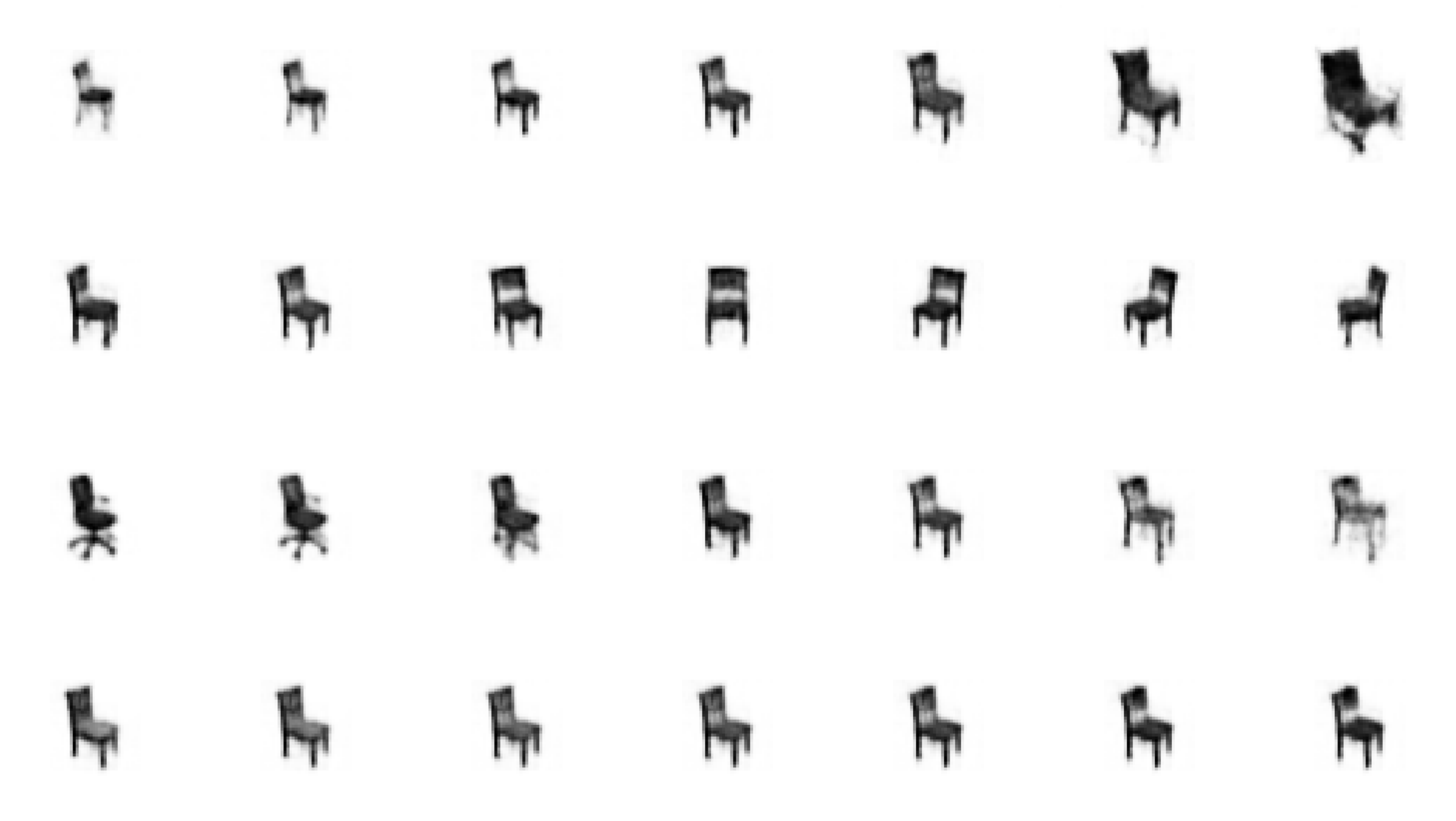}
        \centering\small{Stage (3)}\\
    \end{minipage}%
    }%
        \caption{\textbf{FVAE disentangles action sequences step-by-step.} Latent traversals at each stage. The left is the results on dSprites. The right is the results on 3D Chairs}
        \label{fig:latent_stages}
\end{figure*}

\section{Conclusion}
We demonstrated an example of the effects of images' orientation on the disentangled representations.
We have further investigated the inductive biases on the data by introducing the concept of disentangling action sequences, and we regarded that as discovering the commonality between the things, which is essential for disentanglement.
The experimental results revealed that the actions with higher significance have a larger value of thresholds of latent information.
We further proposed the fractional variational autoencoder (FVAE) to disentangle the action sequences with different significance step-by-step.
We then evaluated the performance of FVAE on dSprites and 3D Chairs.
The results suggested robust disentanglement where re-entangling is prevented.

This paper proposed a novel tool to study the inductive biases by action sequences.
However, other properties of inductive biases on the data remain to be exploited.
The current work focuses on an alternative explanation for disentanglement from the perspective of information theory. In the future, the influence of independence on disentanglement requires further investigation.
% Furthermore, this work did not optimize objective for the FVAE on account of simplification and focalization. This consideration endows the FAVE with a potential to be further improved.

% \section*{Acknowledgements}

% \bibliography{iclr2021_conference}
\bibliography{disentanglement}
\bibliographystyle{iclr2021_conference}

\appendix

\section{Appendix}
\subsection{Datasets}
\subsection{dSprites and 3D Chairs}
\begin{figure}[h]
    \centering
    \subfigure[]{\includegraphics[width=0.45\linewidth]{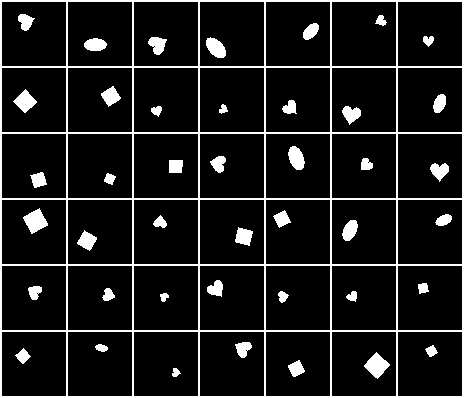}}
    \subfigure[]{\includegraphics[width=0.45\linewidth]{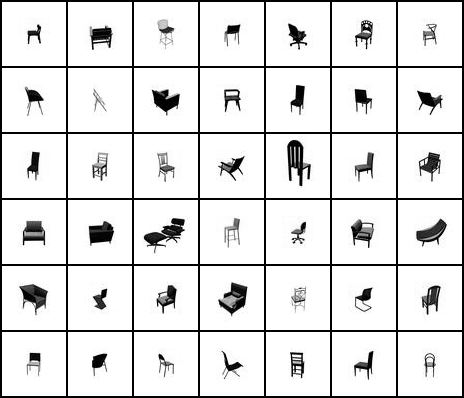}}
    \caption{Real samples from the training dataset.}
\end{figure}

\subsubsection{Designed Datasets}

% \begin{figure}[h]
% \centering
%     \subfigure[A1]{
%     \includegraphics[width=.25\linewidth]{pics/A1}
%     }
%     \subfigure[A2]{
%     \includegraphics[width=.25\linewidth]{pics/A2.pdf}
%     }
% \centering
% \label{fig:factors}
% \caption{Positions of A1, A3 and A2.}
% \end{figure}
\begin{figure}
\centering
    \subfigure[A1]{
    \includegraphics[width=.3\linewidth]{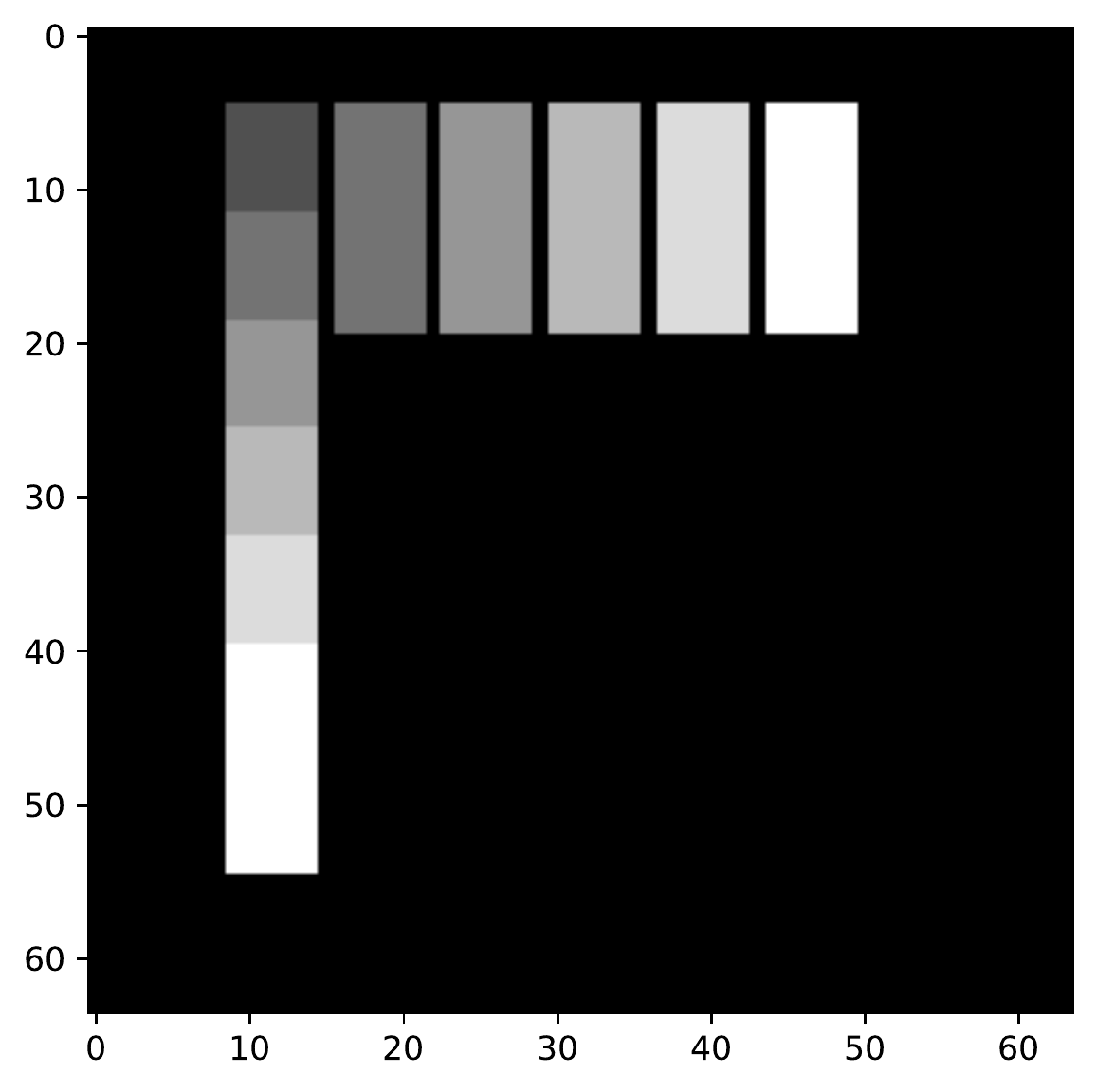}
    }
    \subfigure[A2]{
    \includegraphics[width=.3\linewidth]{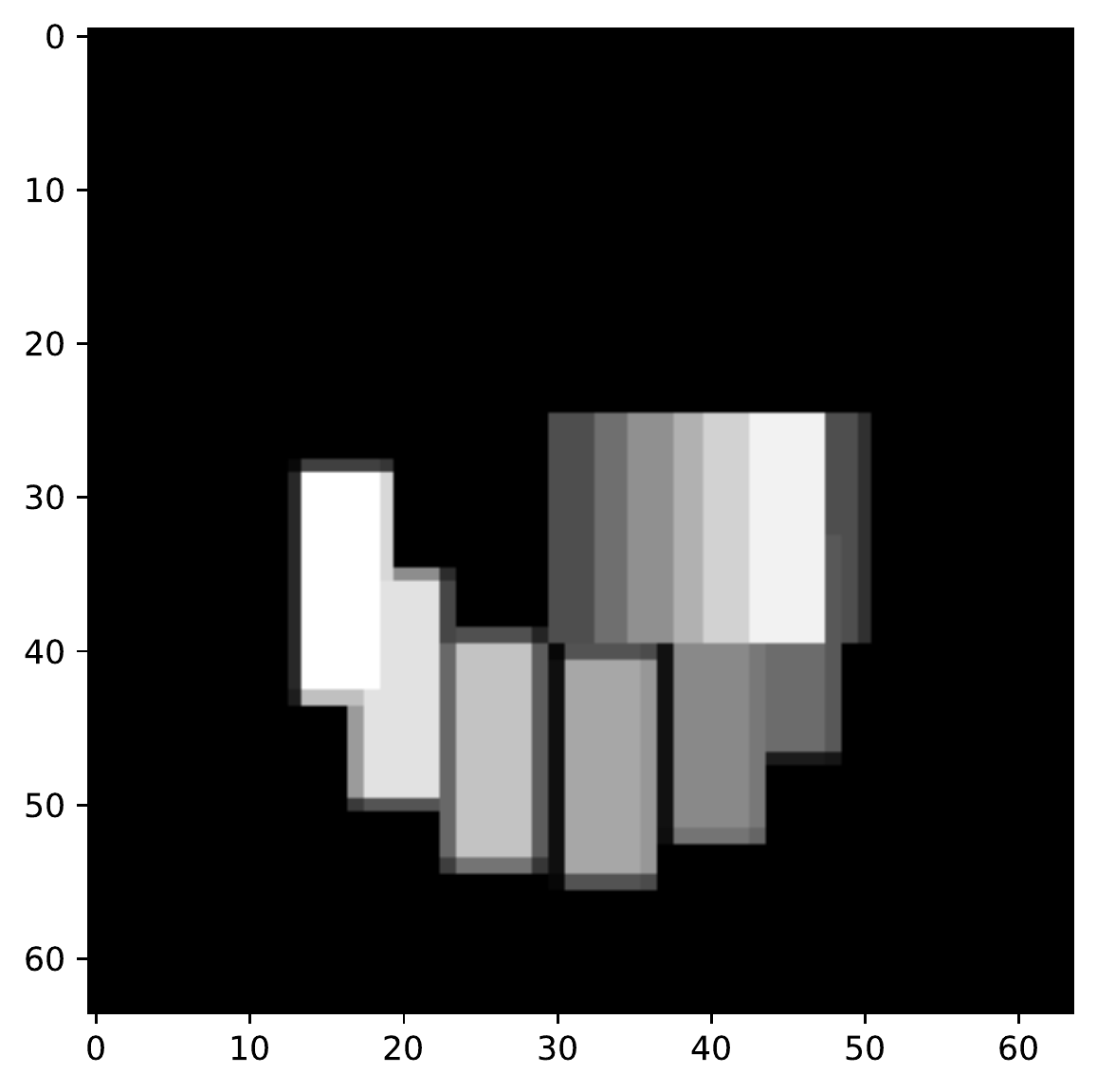}
    }
    \subfigure[A3]{
    \includegraphics[width=.3\linewidth]{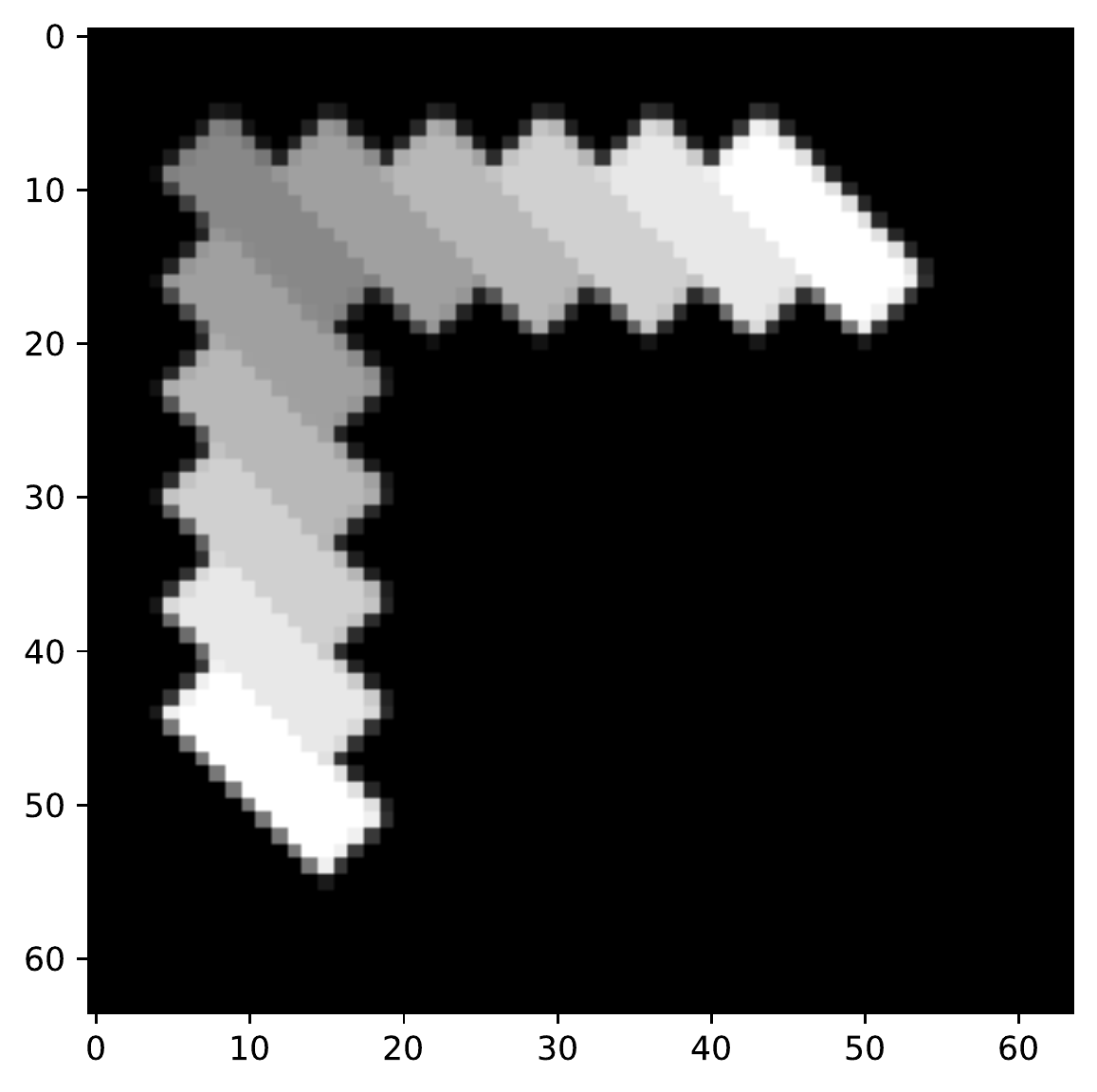}
    }
\caption{Samples on A1, A2, A3. We visualize these samples by afterimage, and the image starts with a dark colour, then, it goes brighter as traversing the factor.}
\label{fig:translating}
\end{figure}

% \begin{figure}
%     \centering
%     \includegraphics[width=.3\linewidth]{pics/A4_translating.pdf}
%     \caption{Caption}
%     \label{fig:my_label}
% \end{figure}

\subsection{Training details}
The basic architecture for all experiments follows the settings on \cite{Burgess.2018}. The hyperparameters of our proposed methods are listed in Tab. \ref{tab:setting_dsprites}. Tab. \ref{tab:threshold_chairs} and \ref{tab:threshold_dsprites} show the measured thresholds of the intrinsic action sequences.
\begin{table}[tb]
    \centering

    \begin{tabular}{|c|c|c|c|}
    \hline
    Phase   & 1             & 2             & 3             \\ \hline
     \hline
    $E_1$     & \textbf{5e-4} & 5e-5          & 5e-5          \\ \hline
    $E_2$     & 0             & \textbf{5e-4} & 5e-5          \\ \hline
    $E_3$     & 0             & 0             & \textbf{5e-4} \\ \hline
     \hline
    $\beta_{1}$ & 100           & 40            & 4             \\ \hline
    $\beta_2$ & 60           & 20            & 2             \\ \hline
    \end{tabular}
    
    \caption{Training settings on dSprites and 3D Chairs. $E_i$ denotes the learning rate of $i$-th group of sub-encoder. $\beta$ denotes the regularisation coefficient before the KL divergence, $\beta_1$ for dSprites, $\beta_2$ for 3D Chairs}
    \label{tab:setting_dsprites}
    
\end{table}

\begin{table}
\centering
\begin{minipage}[t]{0.45\linewidth}
\centering
    \begin{tabular}{|c|c|}
    \hline
      Factor      & Threshold \\ \hline
      Shape       & 32        \\ \hline
      Scale       & 100       \\ \hline
      Orientation & 5         \\ \hline
      Position  X & 120+      \\ \hline
      Position  Y & 120+      \\ \hline
    \end{tabular}
    \caption{Thresholds of actions on dSprites.}
    \label{tab:threshold_dsprites}
\end{minipage}
    \begin{minipage}[t]{0.45\linewidth}
    \centering
    \begin{tabular}{|c|c|}
    \hline
     Factor      & Threshold \\ \hline
     Chair size  & 60        \\ \hline
     Leg style  & 20       \\ \hline
     Swivel & 2         \\ \hline
     Unknown & 2         \\ \hline
    \end{tabular}
    \caption{Thresholds of actions on 3D Chairs.}
    \label{tab:threshold_chairs}
\end{minipage}
\end{table}

\subsection{Samples}
\begin{figure}
    \centering
    \includegraphics[width=\linewidth]{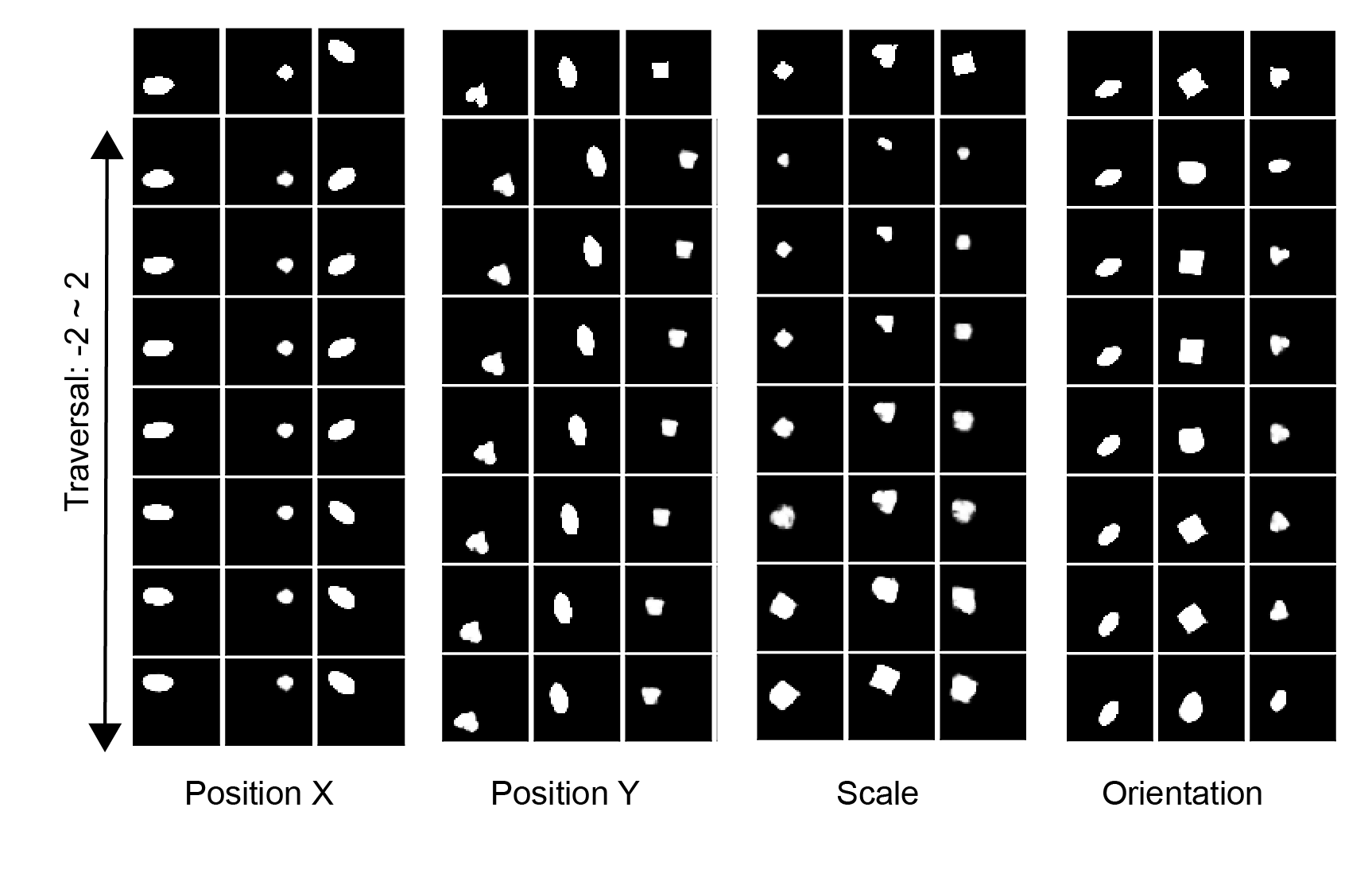}
    \caption{Latent traversal plots for FVAE on dSprites.
    The top row show the real samples from dSprites.
}
\end{figure}

\begin{figure}
    \centering
    \includegraphics[width=\linewidth]{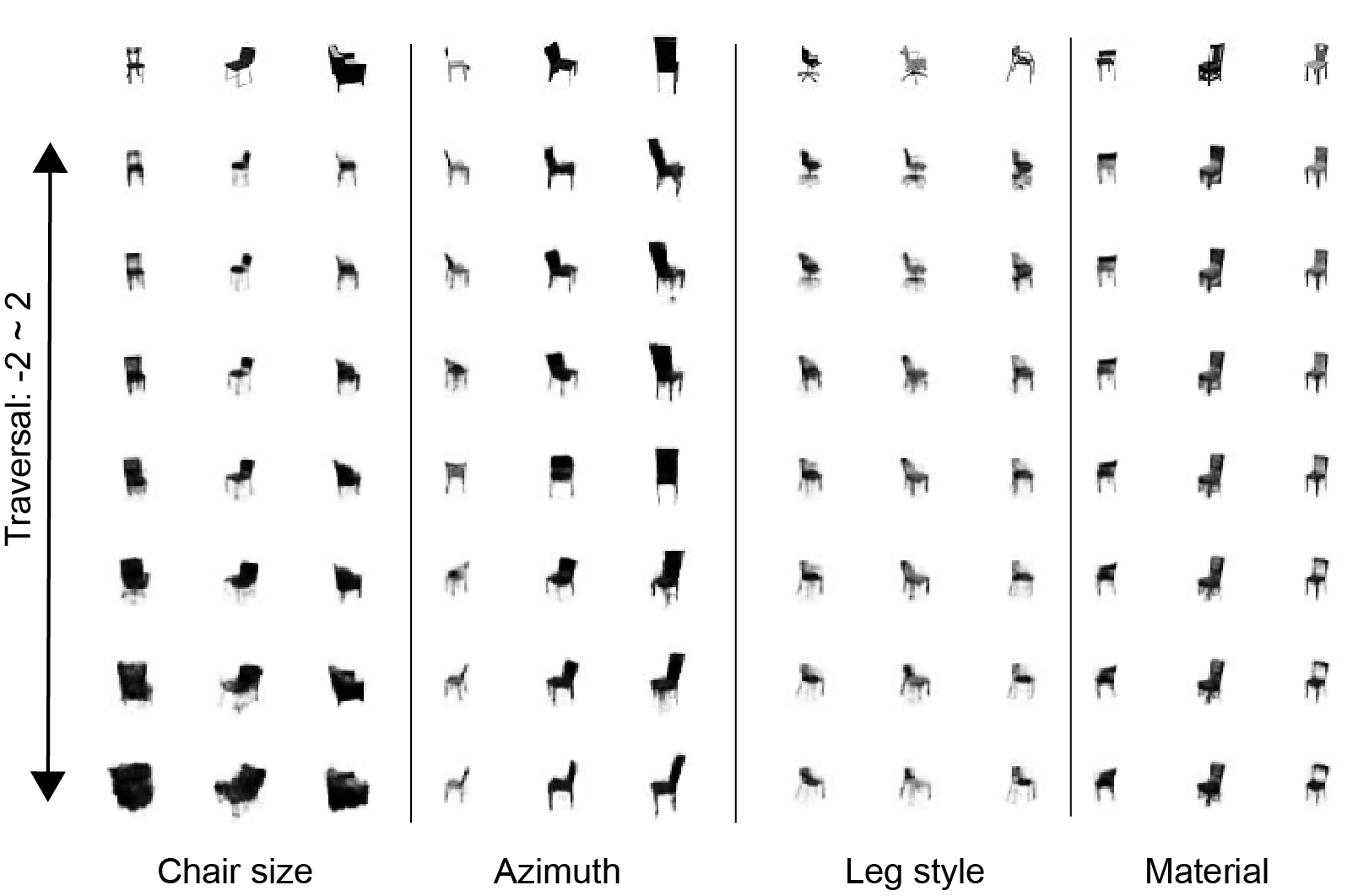}
    \caption{Latent traversal plots for FVAE on 3D Chairs. The top row show the real samples from 3D Chairs.
}
\end{figure}

\end{document}